\documentclass{mva_style}
\usepackage{graphicx}
\usepackage{color}
\usepackage{multirow}
\usepackage{xcolor}
\usepackage{amsmath}

\usepackage{cite}

\usepackage{fix-cm}

\finalcopy 

\usepackage{url}
\begin{document}
\title{Multi-Person Pose Estimation Evaluation \\Using Optimal Transportation and Improved Pose Matching}

\author{
  Takato Moriki ~~~ Hiromu Taketsugu ~~~ Norimichi Ukita\\
  Toyota Technological Institute\\
  {\tt \{25440,25502,ukita\}@toyota-ti.ac.jp}\\
}

\maketitle

\section*{\centering Abstract}
\textit{In Multi-Person Pose Estimation (\textbf{MPPE}), many metrics place importance on ranking of pose detection confidence scores. 
Current metrics tend to disregard false-positive poses with low confidence, focusing primarily on a larger number of high-confidence poses.
Consequently, these metrics may yield high scores even when many false-positive poses with low confidence are detected.
For fair evaluation taking into account a tradeoff between true-positive and false-positive poses, this paper proposes Optimal Correction Cost for pose (\textbf{OCpose}), which evaluates detected poses against pose annotations as an optimal transportation. 
For the fair tradeoff between true-positive and false-positive poses, OCpose equally evaluates all the detected poses regardless of their confidence scores.
In OCpose, on the other hand, the confidence score of each pose is utilized to improve the reliability of matching scores between the estimated pose and pose annotations. 
As a result, OCpose provides a different perspective assessment than other confidence ranking-based metrics.}


\section{Introduction}
\label{section: introduction}

Multi-person pose estimation (MPPE) detects the keypoints of people in images as a set of 2D image coordinates~\cite{survey, survey2, mscoco_kpt, MPII, WAF}.
MPPE is crucial in various applications~\cite{healthcare, sports, security, rehabilitation, Robot}.
MPPE is evaluated based on the rank of the confidence scores of the detected poses in major metrics, such as mAP~\cite{mscoco}, BBP~\cite{BBP}, and others~\cite{PoseTrack}.

\begin{figure}[t]
\centering
    \includegraphics[width=\linewidth]{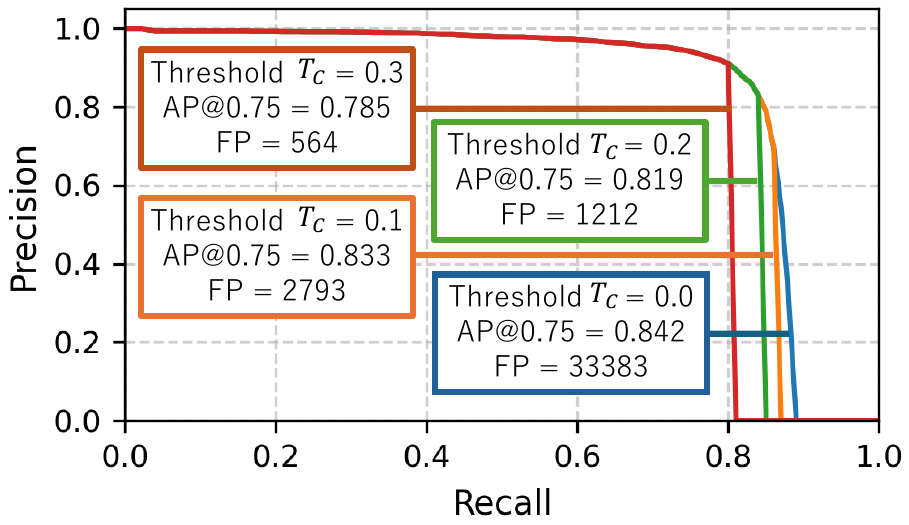}
  \caption{Unfair evaluation in mAP, which is a major evaluation metric using the confidence rank. Lower thresholds lead to high scores (AP) while producing many false-positives (FP).}
  \label{fig: teaser}
\vspace{-20pt}
\end{figure}

While these metrics are widely used in the literature, they have a critical issue in evaluating how each pose estimation method is useful in real applications.
This issue is shown in Fig.~\ref{fig: teaser}, which shows the precision-recall curves drawn by different pose detection thresholds of the same pose estimation method.
Each curve is drawn by calculating the precision score (i.e., the vertical axis) of detected poses selected in descending order of the detection confidence score.
The curve moves to the right as the recall score is incremented along the horizontal axis.
Since many poses are detected with high confidence scores thanks to recent object detection and human pose estimation methods, this curve is almost saturated until the recall score gets larger.
Therefore, in general, the difference between the curves appears after they begin to decline.
Each curve vertically drops to 0 when all detected poses are evaluated.
As shown in Fig.~\ref{fig: teaser}, the curves drop in descending order of the confidence threshold (i.e., $T_{c} =$ 0.3, 0.2, 0.1, and 0.0 in Fig.~\ref{fig: teaser}).
While the area under curve is regarded as the evaluation score (i.e., Average Precision, AP) of each pose estimation method, the scores of these different confidence thresholds are similar because the left area covered by the almost saturated precision scores is dominant.
What is worse is that the evaluation score gradually increases by lowering the confidence threshold even after the curve begins to decline, while the lower threshold produces a large number of false-positve poses.
For example, in Fig.~\ref{fig: teaser}, while the AP score increases from $0.785$ ($T_{c} = 0.3$) to $0.842$ ($T_{c} = 0.0$), the number of false positives significantly increases from $564$ ($T_{c} = 0.3$) to $33,383$ ($T_{c} = 0.0$).

\begin{figure}[t]
\centering
    \includegraphics[width=\linewidth]{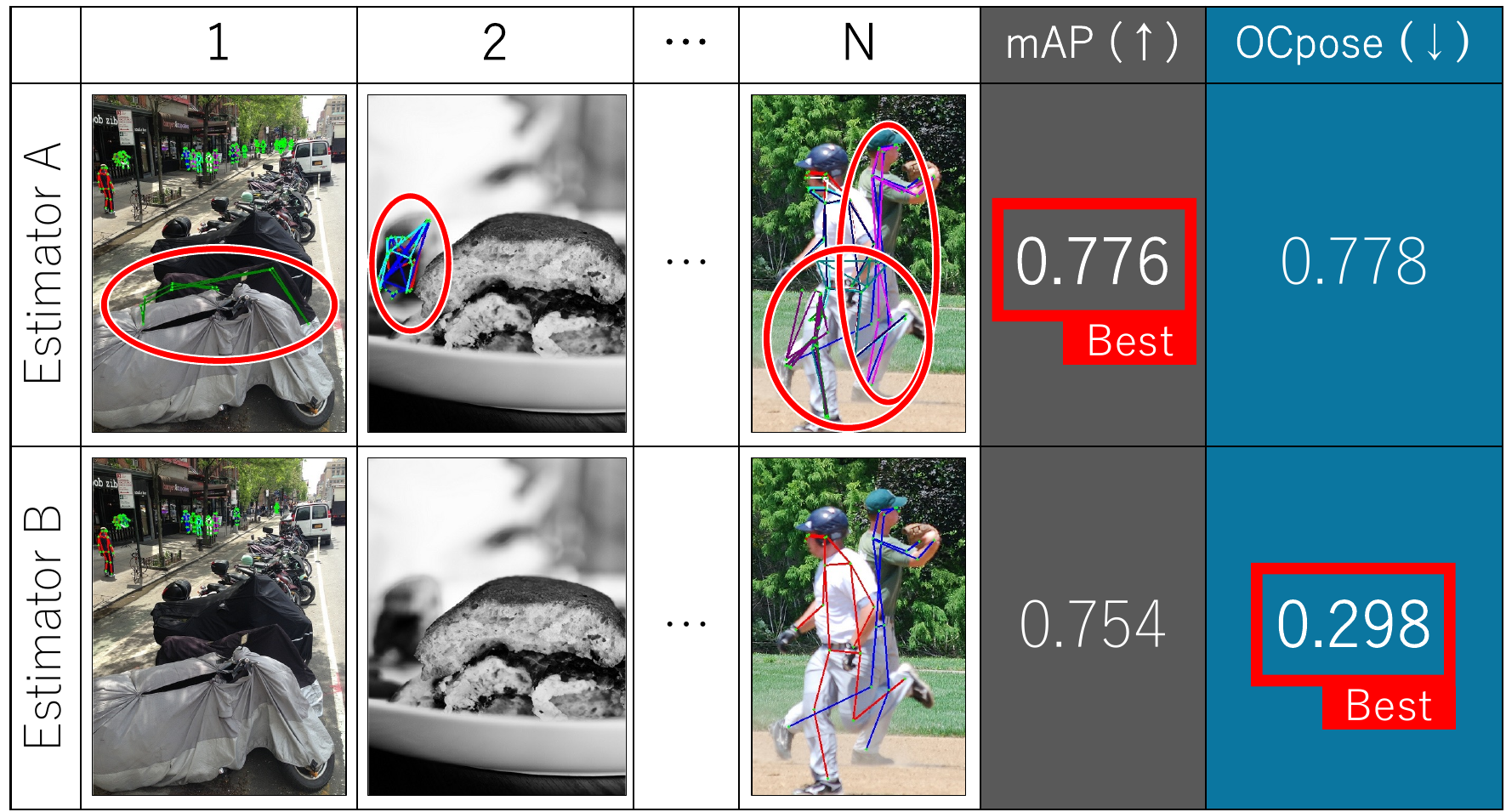}
  \caption{Difference between mAP and OCpose. Red circles highlight false positives. mAP favors Estimator A despite many false positives, whereas OCpose prioritizes Estimator B with fewer false detections.}
  \label{fig: mAP_OCP}
\vspace{-10pt}
\end{figure}

To address the problem of the confidence ranking-based metrics, this paper proposes Optimal Correction Cost for pose (OCpose) with the contributions below:
\begin{itemize}
\setlength{\itemsep}{0pt}
\setlength{\parskip}{0pt}
\setlength{\parsep}{0pt}
    \item \textbf{Evaluation metric without confidence ranking:} OCpose uses optimal transportation (OT) to equally penalize false-positive poses~\cite{OC-cost, OTA, RFPose_OT, Occ_pose, kpt_guide_OT}.
    \item \textbf{Confidence-based pose matching:}     Matching scores between estimated poses and pose annotations are improved with the confidence scores of pose keypoints.
    \item \textbf{User evaluation:} OCpose aligns better with human preferences than other metrics; see Fig.~\ref{fig: mAP_OCP}.
\end{itemize}

\begin{figure}[t]
\noindent
\centering
  \includegraphics[width=\linewidth]{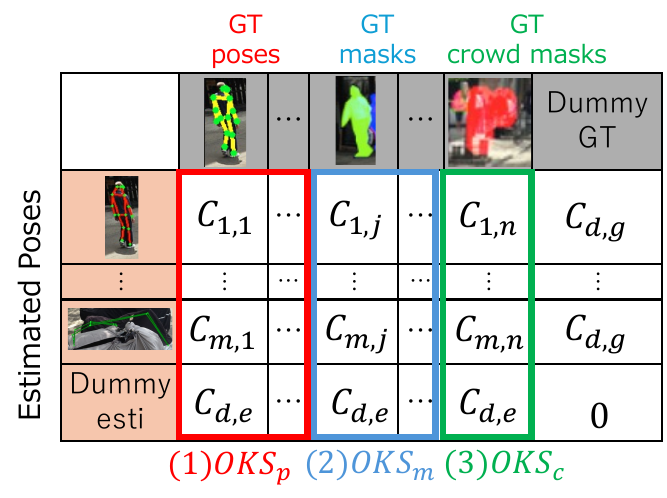}
  \caption{Overview of OCpose. Given GT poses, masks, crowd masks, and estimated poses, OT costs between them are calculated to determine the evaluation score.}
  \label{fig: ocpose}
\vspace{-10pt}
\end{figure}


\section{Related Work}
\label{section: related}

\subsection{Multi-Person Pose Estimation Evaluation}
\label{subsection: related_mppe}

In MPPE, pose similarity between each estimated pose and a Ground Truth (GT) annotation pose is used to evaluate the pose estimation score of the whole image.
Note that this similarity differs from the pose detection confidence given by a pose estimation network; after each pose with its confidence is detected, the pose similarity between the detected pose and each GT pose is evaluated.
For example, Object Keypoint Similarity~\cite{mscoco_kpt} (\textbf{OKS}) is used as the pose-matching scores in mAP.
OKS is calculated by the distance between the keypoints of an estimated pose and its nearest GT.
If OKS exceeds a threshold, the estimated pose is considered a true positive; otherwise, it is a false positive.

However, a major issue in MPPE is that GT poses are unavailable for several people, such as tiny people.
To address this issue, other simpler annotations are provided.
For example, in The Microsoft Common Objects in Context 2017 (\textbf{COCO})~\cite{mscoco}, people without GT pose annotations are with either of ``GT bounding box (bbox)'' given to each person or ``GT crowd bbox'' given to a person crowd.
As with the GT pose, these GT bbox and GT crowd bbox are used to calculate OKS of each estimated pose.
In several metrics~\cite{BBP, PoseTrack}, if OKS exceeds the threshold, this pose is excluded from the evaluation. Otherwise, this pose is regarded as a false positive.
However, these metrics rely on confidence ranking, so they do not address the problem highlighted in Sec.~\ref{section: introduction}.

\subsection{Optimal Transportation Evaluation}
\label{subsection: related_ot}

OT computes a distance between probability distributions.
In~\cite{OTA}, OT is incorporated into the object detection task.
OC-cost~\cite{OC-cost} uses OT as a metric for object detection.
Since OT evaluates detection accuracy and overdetectability independently of the rank of detection confidences, our OCpose also employs OT.


\section{Proposed Evaluation Metric for MPPE}

OCpose calculates the score in two steps. The first step (Sec.~\ref{subsection: matching}) is to calculate the matching scores of all possible combinations between estimated poses and GT annotations (Fig.~\ref{fig: ocpose}).
The second step (Sec.~\ref{subsection: optimization}) is the combinatorial optimization of the matching scores. 

\subsection{Pose Matching}
\label{subsection: matching}

In OCpose, the following three matching scores (OKS$_{p}$, OKS$_{m}$, and OKS$_{c}$) are used.

\textbf{Matching with GT poses.}
OKS$_{p}$, which is the matching score between an estimated pose and a GT pose (shown in Fig.~\ref{fig: ocpose} (1)), is defined by the 2D keypoint coordinates of these two poses~\cite{mscoco}; as the distance between the keypoints of these two poses gets smaller, OKS$_{p}$ becomes larger.
The reliability of OKS$_{p}$ is improved by the visibility of each GT keypoint, which is available as the annotation, so that invisible keypoints are not used.
OCpose uses OKS$_{p}$ as is.

\begin{figure}[t]
\centering
  \includegraphics[width=\linewidth]{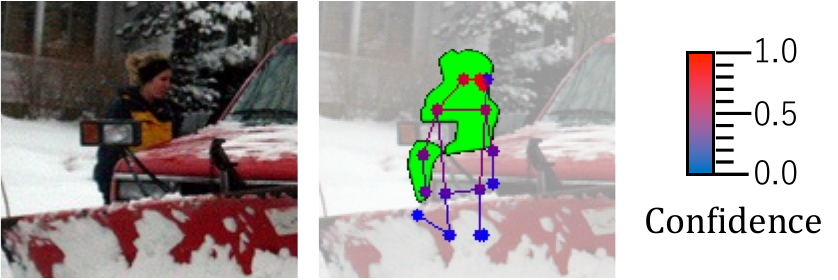}
  \hspace*{6mm} (a) image \hspace*{10mm}
  (b) keypoints \hspace*{28mm}
  \caption{Confidence-based pose matching. Although several keypoints are located outside the mask, their low confidence reduces the influence on the OKS. In contrast, conventional OKS may penalize such poses regardless of confidence.}
  \label{fig:oks_conf}
  \vspace{-10pt}
\end{figure}

\textbf{Matching with GT masks.}
In the literature, OKS$_{m}$, originally proposed for evaluating object detections, is reused as the matching score between an estimated pose and a GT bbox as follows:
\begin{equation}
    \mathrm{OKS_{\textit{b}}} = \frac{1}{N} \sum_{n=1}^{N} \exp \left( -\frac{(d_{b}(x_d^n, y_d^n, b))^2}{2s^2k_n^2} \right)
\label{eq: oks_bbox}
\end{equation}
\begin{equation}
    d_{\textit{b}}(x_d^n, y_d^n, b) = \min(x_d^n, y_d^n, b)
\label{eq: d_bbox}
\end{equation}
where $\min(x_d^n, y_d^n, b)$ is the minimum distance from the estimated keypoint $(x_d^n, y_d^n)$ to its nearest boundary of the GT bbox $b$.
Since the bbox is larger than the set of GT keypoints, Eq.~(\ref{eq: oks_bbox}) over-accepts the detected poses as true positives.
While Eq.~(\ref{eq: oks_bbox}) uses the GT bbox because the prediction output of object detection is a bbox, our matching function should be optimized for human keypoints.
Thus, OCpose replaces a GT bbox in Eq.~(\ref{eq: oks_bbox}) with a pixel-wise mask denoted by \textit{m}:
\begin{equation}
    \mathrm{OKS_{\textit{m}}} = \frac{1}{N} \sum_{n=1}^{N} \exp \left( -\frac{(d_{m}(x_d^n, y_d^n, c_d^n, \textit{m}))^2}{2s^2k_n^2} \right)
\label{eq: oks_mask}
\end{equation}
\begin{equation}
    d_{\textit{m}}(x_d^n, y_d^n, c_d^n, \textit{m}) = \min(x_d^n, y_d^n, \textit{m}) \times \frac{c_d^n}{\sum_{n=1}^{N} c_d^n},
\label{eq: d_mask}
\end{equation}
where $\min(x_d^n, y_d^n, m)$ is the minimum distance from
$(x_d^n, y_d^n)$ to its nearest pixel on the mask. 
$c_d^n$ denotes the confidence score of the $n$-th keypoint of a detected person.
$c_d^n$ encourages OKS$_{m}$ to be larger with keypoints with higher confidence, each of which is expected to be visible inside the mask (Fig.~\ref{fig:oks_conf}).
Therefore, OKS$_{m}$ can avoid the incorrect over-acceptance of true positives.

\begin{figure}[t]
\noindent
\centering
  \includegraphics[width=\linewidth]{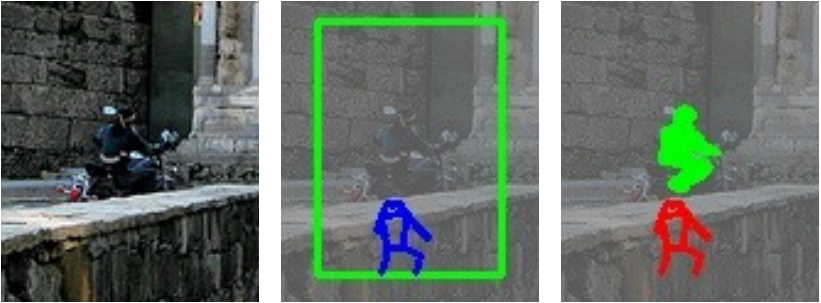}
  (a) image \hspace*{12mm}
  (b) bbox \hspace*{12mm}
  (c) mask
  \caption{Comparison of bbox and mask annotations, which are indicated by green rectangles in (b) and green pixels in (c), respectively. While the bbox incorrectly accepts the false pose colored blue in (b), the mask can recognize it as a false positive in (c).}
  \label{fig: bbox_mask}
\vspace{-10pt}
\end{figure}

\textbf{Matching with GT crowd masks}
OKS$_{c}$ is the matching score between an estimated pose and a GT crowd mask (shown in Fig.~\ref{fig: ocpose} (3)).
The definition and motivation of OKS$_{c}$ are the same as those of OKS$_{m}$.
That is, to suppress false positives detected due to a large number of non-human pixels in a crowd bbox, OCpose uses pixel-wise crowd masks.

\subsection{Combinatorial Optimization}
\label{subsection: optimization}

In Eq.~(\ref{eq: cost}), the matching scores ($OKS_p$, $OKS_m$, $OKS_c$) are used to compute the OT costs $C_{i,j}$ stored in a table, which is called the cost matrix shown in Fig.~\ref{fig: ocpose}.
\begin{equation}
    \mathrm{C}(i, j) = 1 - \mathrm{OKS}(d_{i}, g_{j}),
    \label{eq: cost}
\end{equation}
where $OKS(d_{i}, g_{j})$ is OKS between the $i$-th estimated pose $d_{i}$ and the $j$-th GT annotation $g_{j}$, including GT poses, GT bboxes, and GT crowd bboxes.
The range of $\mathrm{C}(i, j)$ is $0 \leq \mathrm{C}(i, j) \leq 1$.
While OC-cost~\cite{OC-cost} for object detection evaluation has only GT bboxes, corresponding to GT poses for MPPE evaluation, our OCpose has GT masks and GT crowd masks as well.

OT optimizes the combination between the estimated poses and the GT annotation so that each estimated pose matches with one of the GT annotation by minimizing the following function:
\begin{equation}
    \mathrm{OC_{pose}} = \frac{1}{| \Pi_{1} |} \sum_{i=1}^{N_{g}} \sum_{j=1}^{N_{e}} \mathrm{C}(i, j) \cdot \pi_{i,j},
    \label{eq: ocpose}
\end{equation}
where $N_{e}$ and $N_{g}$ denote the numbers of the estimated poses and the GT annotation, respectively.
In Eq.~(\ref{eq: ocpose}), $\pi_{i,j} \in \{ 0, 1 \}$ is optimized. If $\pi_{i,j} = 1$, the $i$-th estimated pose matches with the $j$-th GT annotation.
$\Pi_{1}$ is defined so that $\pi_{i,j}$ is included in $\Pi_{1}$, if $\pi_{i,j}=1$.
While each of the GT poses and the GT masks matches with only one estimated pose, each GT crowd mask can match with multiple estimated poses.
This is the difference from OC-cost~\cite{OC-cost} because OC-cost has no GT crowd masks.
On the other hand, similar to OC-cost, our OCpose also implements dummy GTs to match over-detections when $N_{e} > N_{g}$, which imposes the costs. Also, when $N_{e} < N_{g}$, dummy estimated poses are used to impose costs. OCpose sets these costs to 1.

\section{Experiments}

\subsection{Datasets}
We evaluate OCpose using two standard benchmarks for MPPE. The first one is COCO~\cite{mscoco}, which provides annotations for both MPPE and instance segmentation masks. Another one is CrowdPose Dataset~\cite{crowdpose}, which is specifically designed to evaluate MPPE in crowded scenes. Since CrowdPose Dataset does not provide mask annotation, we generated pseudo masks by state-of-the-art segmentation models~\cite{SAM, Dinov2} using the provided GT bboxes.

\begin{figure*}[t]
\noindent
\centering
  \includegraphics[width=\linewidth]{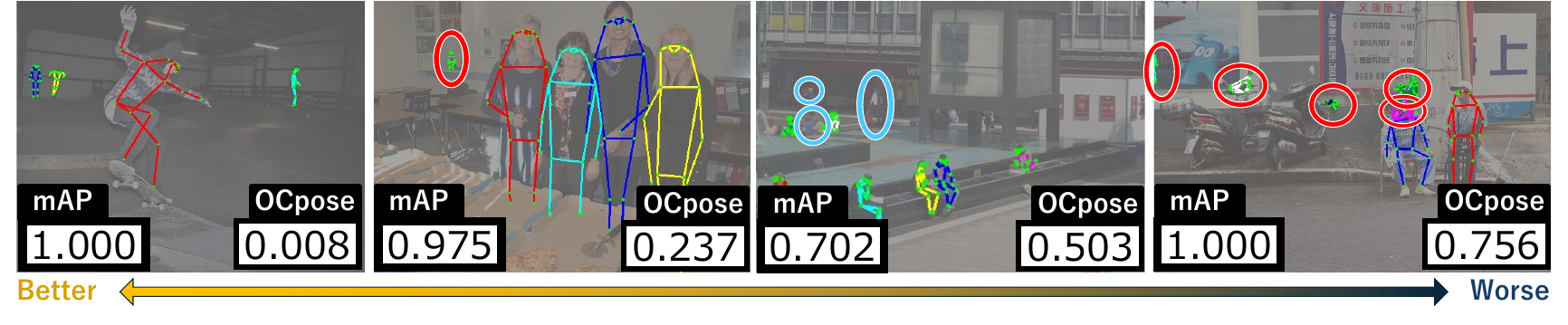}
  \caption{Example of OCpose. The score is descending from left to right. The leftmost image succeeds in accurate MPPE. Other images have missing poses (blue circles) or false positives (red circles).}
  \label{fig:example_OCpose}
\end{figure*}

\begin{table}[t]
    \caption{Comparison of evaluations in COCO. ``threshold'' indicates confidence thresholds where ``-'' represents the one optimized by the authors of each paper and ``o'' represents the one minimizing OCpose.}
    \begin{center}
        \begin{tabular}{llcc}
        \hline
        & threshold & mAP($\uparrow$) & OC pose($\downarrow$) \\
        \hline
        \multirow{2}{*}{BUCTD\cite{BUCTD}} & - (0.00) & \textcolor{red}{0.776} & 0.778 \\
        & o (0.16) & 0.761 & \textcolor{red}{0.292} \\
        \multirow{2}{*}{RTMO\cite{RTMO}} & - (0.10) & \textcolor{red}{0.724} & 0.621 \\
        & o (0.45) & 0.706 & \textcolor{red}{0.381} \\
        \multirow{2}{*}{CID\cite{CID}} & - (0.01) & \textcolor{red}{0.694} & 0.537 \\
        & o (0.04) & 0.668 & \textcolor{red}{0.403} \\
        \multirow{2}{*}{ViTPose\cite{ViTPose}} & - (0.00) & \textcolor{red}{0.768} & 0.792 \\
        & o (0.41) & 0.748 & \textcolor{red}{0.285} \\
        \multirow{2}{*}{HRNet\cite{HRNet}} & - (0.00) & \textcolor{red}{0.763} & 0.802 \\
        & o (0.34) & 0.748 & \textcolor{red}{0.288} \\
        \hline
        \end{tabular}
        \label{tab:mAP_vs_OCP}
    \end{center}
    \vspace{-10pt}
\end{table}

\begin{table}[t]
    \caption{Comparison of evaluations in CrowdPose. ViTPose and HRNet are not shown because human pose detections are not officially available.}
    \begin{center}
        \begin{tabular}{llcc}
        \hline
        & threshold & mAP($\uparrow$) & OCpose($\downarrow$) \\
        \hline
        \multirow{2}{*}{BUCTD\cite{BUCTD}} & - (0.00) & \textcolor{red}{0.785} & 0.891 \\
        & o (0.18) & 0.744 & \textcolor{red}{0.239} \\
        \multirow{2}{*}{RTMO\cite{RTMO}} & - (0.10) & \textcolor{red}{0.839} & 0.412 \\
        & o (0.46) & 0.824 & \textcolor{red}{0.188} \\
        \multirow{2}{*}{CID\cite{CID}} & - (0.01) & \textcolor{red}{0.712} & 0.308 \\
        & o (0.02) & 0.703 & \textcolor{red}{0.255} \\
        \hline
        \end{tabular}
        \label{tab:mAP_vs_OCP_crowdpose}
    \end{center}
    \vspace{-10pt}
\end{table}

\subsection{Quantitative Results}
\label{subsection: quantitative results}
We evaluate state-of-the-art MPPE methods~\cite{BUCTD, RTMO, CID, ViTPose, HRNet}, using both their default confidence thresholds and new thresholds optimized to minimize the OCpose score (Tables~\ref{tab:mAP_vs_OCP}, \ref{tab:mAP_vs_OCP_crowdpose}).
We show that optimizing for OCpose leaves mAP nearly unchanged while significantly reducing the OCpose score, and that optimal thresholds vary substantially across models.
This optimizing is demonstrated in Sec.~\ref{subsection: qualitative results}, \ref{subsection: Agreement with Human Preference}.

\subsection{Qualitative Results}
\label{subsection: qualitative results}
Examples of OCpose (Fig.~\ref{fig:example_OCpose}) show that OCpose becomes lower (i.e., better) if only true-positive poses are detected, as shown in the leftmost example.
We can see that OCpose appropriately penalizes not only false-negatives (i.e., blue circles in Fig.~\ref{fig:example_OCpose}) but also false-positives (i.e., red circles in Fig.~\ref{fig:example_OCpose}).

\subsection{Agreement with Human Preference}
\label{subsection: Agreement with Human Preference}
\begin{figure}[t]
\centering
    \includegraphics[width=\linewidth]{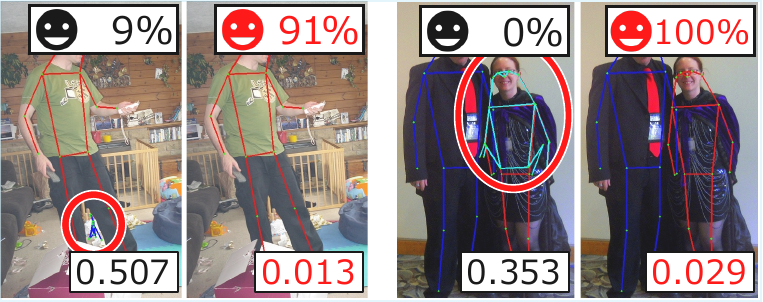}
  \caption{Example images of subjective evaluations. In each image pair, the left one is the result with the default threshold, and the right one is the result with the threshold optimized by OCpose. The percentage next to the smile button represents the user preference. The OCpose scores are in the lower right corner. Red circles indicate false positives. The trends of OCpose and the user preference are consistent.}
  \label{fig:user}
  \vspace{-10pt}
\end{figure}

In our subjective evaluation with 36 participants, human poses detected by CID~\cite{CID} with its default confidence threshold are compared with those detected by the OCpose-optimized threshold. Subjects selected the better one from these two poses detected in 100 randomly selected COCO images (Fig.~\ref{fig:user}).
The results of the OCpose-optimized threshold were selected with \textbf{83.3} percent of the evaluations.

\section{Conclusion}
This paper proposes OCpose, a novel evaluation metric for MPPE that addresses the problems of mAP.
By the optimal transportation with improved cost calculation, OCpose appropriately evaluates false-positive poses, which mAP tends to neglect.
Our OCpose suggests the benefit of having a different perspective from other evaluation metrics, such as mAP.
We believe this contribution provides a more thorough evaluation framework for MPPE methods and serves as a valuable toolkit for both researchers and downstream application developers.



\bibliographystyle{unsrt}
\bibliography{main}
\end{document}